\documentclass[runningheads,a4paper]{llncs}

\makeatletter
\newcommand{\printfnsymbol}[1]{%
  \textsuperscript{\@fnsymbol{#1}}%
}
\makeatother

\usepackage{amssymb}
\usepackage{amsmath}
\usepackage{graphicx}

\usepackage{url}
\usepackage{xcolor}
\urldef{\mailsa}\path|{yuxing.tang, rms}@nih.gov|
\usepackage{diagbox}
\usepackage{multirow}
\newcommand{\etal}{\textit{et al}. }
\newcommand{\ie}{\textit{i}.\textit{e}.}

\begin{document}

\mainmatter  % start of an individual contribution

\title{Attention-Guided Curriculum Learning for Weakly Supervised Classification and Localization of Thoracic Diseases on Chest Radiographs}

\titlerunning{Attention-Guided Curriculum Learning for Thoracic Disease Recognition}

\author{Yuxing Tang$^1$
\thanks{Equal contribution}
\and Xiaosong Wang$^1$\printfnsymbol{1} \and Adam P. Harrison$^1$ \and Le Lu$^1$ \and Jing Xiao$^2$ \and Ronald M. Summers$^1$
}

\authorrunning{Tang et al. }
\institute{% Department of Radiology and Imaging Science, Clinical Center,\\
 $^1$National Institutes of Health, Clinical Center, Bethesda, MD, USA\\
 $^2$Ping An Technology Co., Ltd., Shenzhen, China\\
\mailsa}
% \mailsb\\
% \url{http://www.springer.com/lncs}
% }

\toctitle{Lecture Notes in Computer Science}
\tocauthor{Authors' Instructions}
\maketitle

\begin{abstract}
In this work, we exploit the task of joint classification and weakly supervised localization of thoracic diseases from chest radiographs, with only image-level disease labels coupled with disease severity-level (DSL) information of a subset. A convolutional neural network (CNN) based attention-guided curriculum learning (AGCL) framework is presented, which leverages the severity-level attributes mined from radiology reports. Images in order of difficulty (grouped by different severity-levels) are fed to CNN to boost the learning gradually. In addition, highly confident samples (measured by classification probabilities) and their corresponding class-conditional heatmaps (generated by the CNN) are extracted and further fed into the AGCL framework to guide the learning of more distinctive convolutional features in the next iteration. A two-path network architecture is designed to regress the heatmaps from selected seed samples in addition to the original classification task. The joint learning scheme can improve the classification and localization performance along with more seed samples for the next iteration. We demonstrate the effectiveness of this iterative refinement framework via extensive experimental evaluations on the publicly available ChestXray14 dataset. AGCL achieves over 5.7\% (averaged over 14 diseases) increase in classification AUC and 7\%/11\% increases in Recall/Precision for the localization task compared to the state of the art.

\end{abstract}

\section{Introduction}
The chest X-ray (radiograph) is a fast and painless screening test that is commonly performed to diagnose various thoracic abnormalities, such as pneumonias, pneumothoraces and lung nodules. It is one of the most cost-effective imaging examinations and imparts minimal radiation exposure to the patient while displaying a wide range of visual diagnostic information. Identifying and distinguishing the various chest abnormalities in chest X-rays is a challenging task even to the human observer. Therefore, its interpretation has been performed mostly by board-certified radiologists or other physicians. There are huge demands on developing computer-aided detection (CADe) methods to assist radiologists and other physicians in reading and comprehending chest X-ray images.

%%%%%%%%%%%%%%%%%%%%%%%%%%%%%%%Fig%%%%%%%%%%%%%%%%%%%%%%%%%%%%%%%%%%%%%%%%%%%%%%%%%%%
\begin{figure*}[t!]
	\centering
	\includegraphics[width=0.8\linewidth]{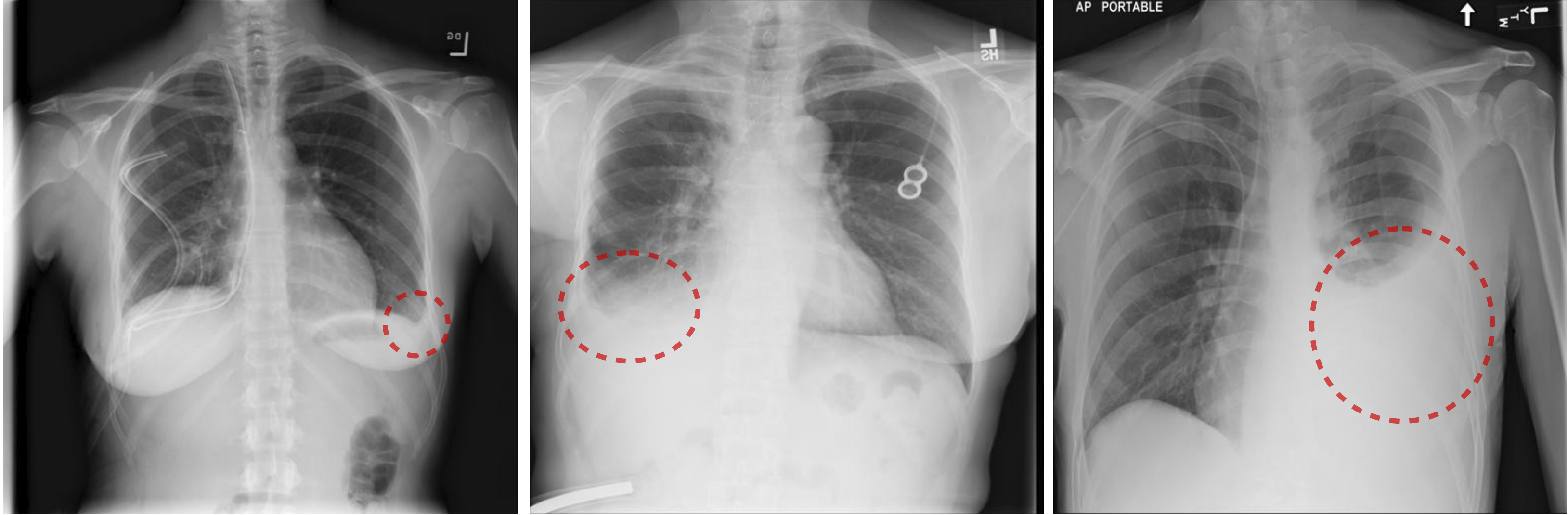}
	\caption{\textbf{Left}: \textit{small} left pleural effusion. \textbf{Middle}: \textit{moderate} right effusion. \textbf{Right}: \textit{large} left pleural effusion.
	}
	\label{fig:severity}
\end{figure*}
%%%%%%%%%%%%%%%%%%%%%%%%%%%%%%%%%%%%%%%%%%%%%%%%%%%%%%%%%%%%%%%%%%%%%%%%%%%%%%%%%%%%%%

Currently, deep learning methods, especially convolutional neural networks (CNN)~\cite{AlexNet_NIPS2012,Tang_CVPR16}, have become ubiquitous. They have achieved compelling performance across a number of tasks in the medical imaging domain~\cite{Yan_JMI_2018,Tang_MICCAI18}. Most of these applications typically involve only one particular type of disease or lesion, such as automated classification of pulmonary tuberculosis~\cite{Lakhani_Radiology_2017}, pneumonia detection~\cite{Rajpurkar_2017_chexnet}, and lung nodule segmentation~\cite{Jin_MICCAI18}. Wang~\etal \cite{Wang_CVPR2017} recently introduced a hospital-scale chest X-ray (ChestX-ray14) dataset containing 112,120 frontal-view X-ray images, with 14 thoracic disease labels text-mined from associated radiology reports using natural language processing (NLP) techniques.
Furthermore, a weakly-supervised CNN based multi-label thoracic disease classification and localization framework was proposed in~\cite{Wang_CVPR2017} using only image-level labels. Li~\etal~\cite{Li_2018_CVPR} presented a unified network that simultaneously improves classification and localization with the help of extra bounding boxes indicating disease location.

In addition to the disease labels that represent the presence or absence of certain disease, we also want to utilize the attributes of those diseases contained in the radiology reports. Disease severity level (DSL) is one of the most critical attributes, since different severity levels are correlated with highly different visual appearances in chest X-rays (see examples in Fig. 1). Radiologists tend to state such disease severity levels (\ie, \textit{[minimal, tiny, small, mild], [middle-size, moderate], [remarkable, large, severe]}, etc.) when describing the findings in chest X-rays. This type of disease attribute information can be exploited to enhance and enrich the accuracy of NLP-mined disease labels, which consequently may facilitate to build more accurate and robust disease classification and localization framework than~\cite{Wang_CVPR2017}. More recently, Wang~\etal~\cite{Wang_2018_CVPR} proposed the TieNet (Text-Image Embedding Network), which was an end-to-end CNN-RNN architecture for learning to embed visual images and text reports for image classification and report generation. However, the disease attributes were not explicitly modeled in the TieNet framework.

In this paper, we propose an attention-guided curriculum learning (AGCL) framework for the task of joint thoracic disease classification and weakly supervised localization, where only image-level disease labels and severity level information of a subset are available. Note, we do not use bounding boxes for training. In AGCL, we use the disease severity level to group the data samples as a means to build the curriculum for \textit{curriculum learning}~\cite{bengio_icml2009}. For each disease category, we begin by learning from severe samples, progressively adding moderate and mild samples as the CNN matures and converges gradually by seeing samples from ``easy'' to ``hard''. The intuition behind curriculum learning is to mimic the common human process of gradual learning, starting from the easiest or obvious samples to harder or more ambiguous ones, which is notably the case for medical students learning to read radiographs. Furthermore, we use the CNN generated disease heatmaps (\textit{visual attention}) of ``confident'' \textit{seed} images to guide the CNN in an iterative training process. The initial seeds are composed of: (1) images of severe and moderate disease level, and (2) images with high classification probability scores from the current CNN classifier. A two-path multi-task learning network architecture is designed to regress the heatmaps from selected seed samples in addition to the original classification task. In each iteration, the joint learning scheme can harvest more seeds of high quality as the network fine-tuning process iterates, resulting in increased guidance and more discriminative CNN for better classification and localization.

We test our proposed method on the public ChestXray14 dataset to evaluate the multi-label disease classification and localization performance. Comprehensive experimental results demonstrate the effectiveness of our framework in acquiring high-quality seeds, and the visual attention generated from seed images are evidently beneficial in guiding the learning procedure to improve both the classification and localization accuracy.

\section{Method}

\subsection{A CNN Based Classification and Localization Framework}
\label{baseline_model}

\begin{figure*}[tbp!]
	\centering
	\includegraphics[width=\linewidth]{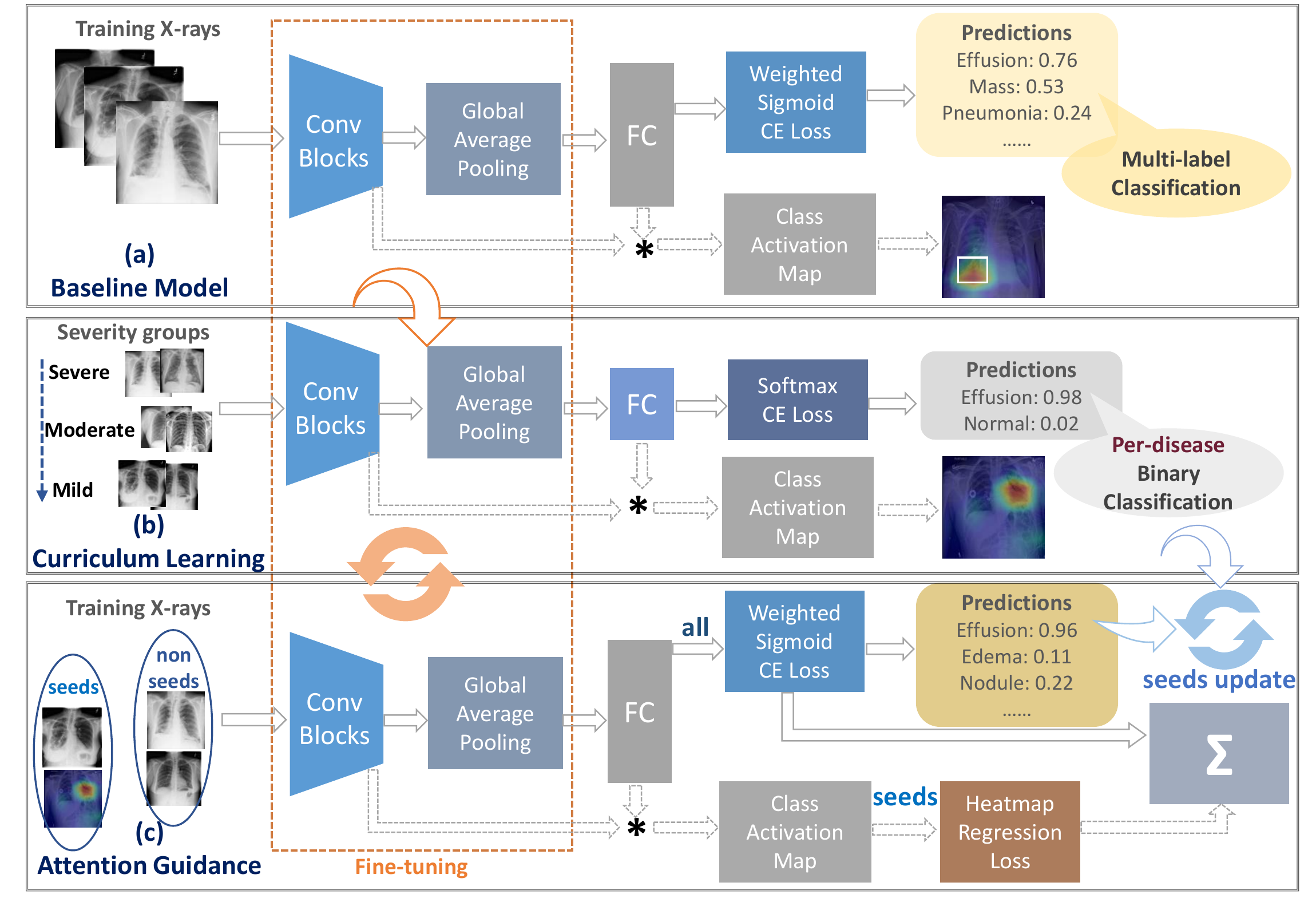}
	\caption{Overall architecture of attention-guided curriculum learning (AGCL).}
	\label{fig:overview-bsl}
\end{figure*}

The proposed AGCL approach starts by initializing a CNN pre-trained from ImageNet and then fine-tuning it on the ChestX-ray14 dataset on all $C$ ($C=14$) disease categories. This is similar to~\cite{Wang_CVPR2017} except that the transition layer is discarded. This serves as the baseline of our method for multi-label classification and localization. 
The flowchart of the baseline framework is shown in Fig.~\ref{fig:overview-bsl}(a). 

\subsection{Disease Severity-Level Based Curriculum Learning}
\label{CL}

Generally, the knowledge to be acquired by students is meticulously designed in a curriculum, so that ``easier'' concepts are introduced first, and more in-depth knowledge is systematically acquired by mastering concepts with increasing difficulty. The ``easy-to-hard'' principle of curriculum learning \cite{bengio_icml2009} has been helpful for both image classification and weakly supervised object detection \cite{Shi_eccv2016} in computer vision. The curriculum that controls what training data should be fed to the model is usually built based on prior information, such as object size (the larger, the easier) or other more sophisticated human supervision.

%%%%%%%%%%%%%%%%%%%%%%%%%%%%%%%Fig%%%%%%%%%%%%%%%%%%%%%%%%%%%%%%%%%%%%%%%%%%%%%%%%%%%
%%%%%%%%%%%%%%%%%%%%%%%%%%%%%%%%%%%%%%%%%%%%%%%%%%%%%%%%%%%%%%%%%%%%%%%%%%%%%%%%%%%%%%

We mine the disease severity level (DSL) attributes from radiology reports using a similar NLP techniques introduced in \cite{Wang_CVPR2017}. Severity descriptions that correlated to the disease keyword are extracted using the dependency graph built on each sentence in the report. DSL attributes are then collected whenever available from the whole training set and are grouped into three clusters, namely \textit{mild}, \textit{moderate} and \textit{severe}. We treat the severity attributes as prior knowledge to build the curriculum. The prediction layers of the baseline model are replaced with a randomly initialized 2-way fully connected (FC) layer and a softmax cross-entropy loss, and we fine-tune the baseline model to a binary classification network for each disease category. The training samples are presented to the network in order of decreasing severity levels (increasing difficulties) of a certain disease, that is from severe samples to moderate to mild gradually as the CNN becomes more adept at later iterations during training. The negative samples come from \textit{normal} cases (without any diseases mentioned in the radiology report) in the dataset and the number of negative samples for each category is balanced with the number of positive samples of that category. Note that we fine-tune from the weights learned from the baseline model in Sec.~\ref{baseline_model} because (1) the images annotated with all severity levels account for only about 25\% of the training samples, which is not sufficient for training a deep CNN with millions of parameters, and (2) the baseline model is expected to have captured an overall concept distribution of the target dataset, which could be a useful starting point for curriculum learning.

The disease-specific class activation map \cite{Zhou_CVPR2016} (CAM, or heatmap) $H^{c}$ of a chest X-ray image for a positive disease class $c$ is:
\begin{equation}\label{eq:cam}
	H^{c}{(x, y)} = \sum_{d=1}^{D}\textbf{w}_{c}(d)*f(x, y, d),
\end{equation}
\noindent where $\textbf{w}_{c}(d)$ is the weights of the FC layer corresponding to the positive disease category $c\in \mathbb{C} = \{1, ..., C\}$ in each binary classification network, and $f(x, y, d)$ is the activation of the $d$-th ($d \in \{1, ..., D\}$) neuron of the last convolutional layer at a spatial coordinate $(x, y)$, where $D$ is number of feature maps.

The heatmaps can be further employed as visual attention guidance for the CNN in the succeeding iterative refinement steps described in Sec. \ref{AGL}. 
The reason to split the network into individual binary models per disease instead of fine-tuning as a whole is that the severity levels tend to be inconsistent among different diseases in a multi-label situation. Moreover, the binary models are empirically found to be more discriminative and spatially accurate on generating disease-specific heatmaps. 

\subsection{Attention Guided Iterative Refinement}
\label{AGL}

In this section, we explore and harvest highly ``confident'' \textit{seed} images. We assume their computed disease-specific heatmaps could highlight the regions that potentially contribute more to the final disease recognition than the \textit{non-seed} images. The highlighted regions represent the Region of Interest (ROI), or in other words, \textit{visual attention} of disease patterns. In addition to the curriculum learning for each disease category shown in Fig.~\ref{fig:overview-bsl}(b), we introduce a heatmap regression path (shown in Fig.~\ref{fig:overview-bsl}(c)) to enforce the attention-guided learning of better convolutional features, which in turn generate more meaningful heatmaps. By using such an iterative refinement loop, we demonstrated that both the classification and localization results could be simultaneously and significantly improved over the baseline.

\textbf{Harvesting Seeds:}
Ideally, image samples with \textit{severe} and \textit{moderate} disease severity levels could potentially be selected as seeds (denoted as $\mathbb{S}_1$) since their visual appearances are relatively easier to recognize than \textit{mild} ones. Additional selection criterion requires that an image is labeled with a certain disease $c$ and is correctly classified by the corresponding binary classifier introduced in Sec.~\ref{CL} with a probability score larger than a threshold $t$ (seeds collected as $\mathbb{S}_2$). We believe that the disease-specific heatmaps (seed attention maps) generated by Eq.~\eqref{eq:cam} from those \textit{seeds} image samples ($\mathbb{S} = \mathbb{S}_1 \cup \mathbb{S}_2$) shall exhibit higher precision in localizing disease patterns than other samples.

\textbf{Attention Guidance from Seeds:}
We create a branch in the original classification network to guide the learning of better convolutional features using the seed attention maps. This branch shares all the convolutional blocks with the baseline model in Sec.~\ref{baseline_model} and includes an additional heatmap regression path. The regression loss is modeled as the sum of the channel-wise smooth $L1$ losses over feature channels between the heatmap generated by the current network ($\hat{H}^{c}$) and the seed attention map of the last iteration ($H^{c}$).

\begin{equation}\label{eq:l1_loss}
	Loss_{reg}(I) = \sum_{d=1}^{D}\sum_{x,y}smooth_{L_1}(\hat{H}^{c}{(x,y)}-H^{c}{(x,y)}), \forall{I}\in\mathbb{S}
\end{equation}
in which

\begin{equation}\label{eq:l1_loss}
	smooth_{L_1}(z) = \left\{
  \begin{array}{l l}
  0.5z^2,&  \quad \text{if $|z|<1$}\\
  |z|-0.5, & \quad \text{otherwise.}
  \end{array} \right.
\end{equation}
% This regression layer is only turned on for images with seed attention maps.
The final objective function to be optimized is a weighted sum of the sigmoid cross-entropy loss for multi-label classification ($Loss_{cls}$) and the heatmap regression loss ($Loss_{reg}$) for localization:
\begin{equation}\label{eq:loss_final}
	Loss_{final}(I) = Loss_{cls}(I) + \lambda \sum_{c=1}^{C}\mathbf{1}_{c} Loss_{reg}(I),
\end{equation}
where $\mathbf{1}_{c}=1$ if an image $I$ is a seed for disease category $c \in C$, otherwise $\mathbf{1}_{c}=0$. $\lambda$ is used to balance the classification and regression loss so that they have roughly equal contributions. We empirically set $\lambda$ to $0.005$. 

\textbf{Harvesting Additional Seeds:}
Once the network is retrained with the attention guidance, the curriculum learning procedures will be conducted again to harvest new confident seeds. All non-seed positive training images in each disease category will be inputted to their corresponding binary classifiers, among which highly scored images are harvested as additional seeds. Together with the initial seeds, their heatmaps are further fed to refinement framework as additional visual attention to guide the CNN to focus on disease-specific attended regions in the chest X-rays. Consequently, more and more confident seeds could be harvested while the accuracy of classification and localization improves gradually.

\section{Experiments}
\label{exp}

We extensively evaluate the proposed AGCL approach on the ChestX-ray14 dataset~\cite{Wang_CVPR2017}, %\footnote{https://nihcc.app.box.com/v/ChestXray-NIHCC},
which contains 112,120 frontal-view chest X-ray images of 30,805 unique patients, with 14 thoracic disease labels, extracted from associated radiology reports using NLP techniques. A small subset of 880 images within the dataset are annotated by board-certified radiologists, resulting in 984 bounding box locations containing 8 types of disease.
We further extracted severity attributes along with the disease keywords. For classification, we use the same patient-level data splits provided in the dataset, which uses roughly 70\% of the images for training, 10\% for validation and 20\% for testing. The disease localization is evaluated on all the 984 bounding boxes (not used for training).

We resize the original 3-channel $1024 \times 1024$ images to $512 \times 512$ pixels due to the trade-off between higher resolution and affordable computational load. ResNet-50 \cite{He_CVPR2016} is employed as the backbone of the proposed CNN architectures. For the baseline method and all the AGCL steps, we optimize the network using SGD with momentum and stop training after the validation loss reaches a plateau. The learning rate is set to be 0.001 and divided by 10 every 10 epochs. The AGCL is implemented using the Caffe framework.

\textbf{Disease Classification:} 
%%%%%%%%%%%%%%%%%%%%%%%%%%%%%%%table%%%%%%%%%%%%%%%%%%%%%%%%%%%%%%%%%%%%%%%%%%%%%%
\begin{table*}[!t] \fontsize{8.5pt}{8.5pt}\selectfont
	\renewcommand{\arraystretch}{1.2}
	\caption{Per-category multi-label classification AUC comparison on the test set of ChestX-ray14. BSL: baseline model. PT: Pleural Thickening. AVG: Average AUC.} 
	\label{table-results-cls-auc}
	\centering
	\begin{tabular}{@{}l||c|c|c|c|c|c|c||c}
		\cline{1-8}
		Disease &Atelectasis &Cardiomegaly &Effusion &Infiltration &Mass &Nodule &Pneumonia\\ \cline{1-8}
		\cite{Wang_CVPR2017} &.7003 &.8100 &.7585 &.6614 &.6933 &.6687 &.6580 \\ \cline{1-8}
        \cite{Wang_2018_CVPR} &.7320 &.8440 &.7930 &.6660 &.7250 &.6850 &.7200 \\ \cline{1-8}
		BSL &.7268 &.8545 &.7899 &.6788 &.7527 &.7143 &.6934 \\ \cline{1-8}
		AGL &.7353 &.8656 &.8113 &.6823 &.7722 &.7178 &.7100 \\ \cline{1-8}
		AGCL-1 &.7500 &.8644 &.8028 &.6748 &.7590 &.7064 &.7130 \\ \cline{1-8}
		AGCL-2 &\textbf{.7557} &\textbf{.8865} &\textbf{.8191} &\textbf{.6892} &\textbf{.8136} &\textbf{.7545} &\textbf{.7292}\\ \hline \hline
		Disease &Pneumothorax &Consolidation &Edema &Emphysema &Fibrosis &PT &Hernia &AVG\\ \hline
		\cite{Wang_CVPR2017} &.7993 &.7032 &.8052 &.8330 &.7859 &.6835 &.8717 &.7451\\ \hline
        \cite{Wang_2018_CVPR} &.8470 &.7010 &.8290 &.8650 &.7960 &.7350 &.8760 &.7724\\ \hline
		BSL &.8260 &.7052 &.8148 &.8698 &.7892 &.7260 &.8500 &.7708\\ \hline
		AGL &.8423 &.7042 &.8366 &.8874 &\textbf{.8180} &.7499 &.8543 &.7777\\ \hline
		AGCL-1 &.8413 &.7209 &.8321 &.8771 &.7922 &.7472 &\textbf{.9012} &.7844\\ \hline
		AGCL-2 &\textbf{.8499} &\textbf{.7283} &\textbf{.8475} &\textbf{.9075} &.8179 &\textbf{.7647} &.8747 & \textbf{.8027}\\ \hline
	\end{tabular}
\end{table*}
%%%%%%%%%%%%%%%%%%%%%%%%%%%%%%%%%%%%%%%%%%%%%%%%%%%%%%%%%%%%%%%%%%%%%%%%%%%%%%%%%%%%%%
We quantitatively evaluate the disease classification performance using the AUC (area under the ROC curve) score for each category. We ablate the curriculum learning step in the AGCL framework to assess its effect. It is denoted by AGL (attention-guided learning), where seeds are only initialized with high scored images from the baseline model. The per-disease AUC comparisons of the benchmark \cite{Wang_CVPR2017}, our baseline method, AGL, and AGCL with one refinement step (AGCL-1) and two refinement steps (AGCL-2) are shown in Table~\ref{table-results-cls-auc}. A higher AUC score implies a better classifier.

Compared with the benchmark results~\cite{Wang_CVPR2017}, our baseline model achieves higher AUC scores for all the categories except \textit{Hernia}, which contains a very limited number ($<200$) of samples. AGL consistently achieves better performance than the baseline model, demonstrating the effectiveness of attention-guided learning in our framework. Furthermore, the proposed AGCL improves upon AGL by introducing more confident heatmaps from seed images using curriculum learning. The iterative refinement process using AGCL is proven to be effective given the fact that AGCL-2 achieves better classification results than AGCL-1. We experimentally find that AGCL-3 has similar results as AGCL-2, which we believe the iterative refinement process has reached the convergence.

%%%%%%%%%%%%%%%%%%%%%%%%%%%%%%%table%%%%%%%%%%%%%%%%%%%%%%%%%%%%%%%%%%%%%%%%%%%%%%
\begin{table*}[t!] \fontsize{8.5pt}{8.5pt}\selectfont
\renewcommand{\arraystretch}{1.2}
\caption{Comparison of disease localization results using $T(IoBB)$ = 0.25. GT: number of ground-truth bounding boxes. BSL: baseline model.}
\label{table-results-loc}
\centering
\begin{tabular}{@{}l|c|c c c|c c c|c c c|c c c}
  \hline
  Disease &GT &\multicolumn{3}{c|}{Detected Box} &\multicolumn{3}{c|}{True Positive} &\multicolumn{3}{c|}{Recall} & \multicolumn{3}{c}{Precision} \\
  & &\tiny{BSL} &\tiny{AGL} &\tiny{\textbf{AGCL}} &\tiny{BSL} &\tiny{AGL} &\tiny{\textbf{AGCL}} &\tiny{BSL} &\tiny{AGL} &\tiny{\textbf{AGCL}} &\tiny{BSL} &\tiny{AGL} &\tiny{\textbf{AGCL}}\\
  \hline
  Atelectasis &180 &734 &687 &\textbf{620} &138 &162&\textbf{190} &0.46 &0.58 &\textbf{0.66} &0.19 &0.24 &\textbf{0.31} \\ \hline
  Cardiomegaly &146 &394 &342 &\textbf{390} &369 &320 &\textbf{366} &0.99 &0.99 &\textbf{1.00} &0.94 &0.94 &\textbf{0.94} \\ \hline
  Effusion &153 &743 &544 &\textbf{546} &205 &218 &\textbf{228} &0.55 &0.64 &\textbf{0.72} &0.28 &0.40 &\textbf{0.42}\\ \hline
  Infiltration &123 &595 &616 &\textbf{501} &221 &232 &\textbf{242} &0.76 &0.80 &\textbf{0.87} &0.42 &0.38 &\textbf{0.48}\\ \hline
  Mass &85 &291 &262 &\textbf{278} &106 &116 &\textbf{148} &0.65 &0.69 &\textbf{0.79} &0.36 &0.44 &\textbf{0.53}\\ \hline 
  Nodule &79 &296 &293&\textbf{295} &16 &28 &\textbf{28} &0.19 &0.32 &\textbf{0.32} &0.05 &0.10 &\textbf{0.09}\\ \hline 
  Pneumonia &120 &558 &535 &\textbf{526} &175 &189 &\textbf{210} &0.76 &0.80 &\textbf{0.82} &0.31 &0.35 &\textbf{0.40}\\ \hline 
  Pneumothorax &98 &394 &359 &\textbf{347} &79 &80 &\textbf{106} &0.45 &0.46 &\textbf{0.49} &0.20 &0.22 &\textbf{0.31}\\ \hline 
  \textbf{Total} &984 &3935 & 3617 &\textbf{3498} &1309 &1345 &\textbf{1518} &0.66 &0.68 &\textbf{0.73} &0.33 &0.37 &\textbf{0.44}\\ \hline
\end{tabular}
\end{table*}
%%%%%%%%%%%%%%%%%%%%%%%%%%%%%%%%%%%%%%%%%%%%%%%%%%%%%%%%%%%%%%%%%%%%%%%%%%%%%%%%%%%%%%
%%%%%%%%%%%%%%%%%%%%%%%%%%%%%%%Fig%%%%%%%%%%%%%%%%%%%%%%%%%%%%%%%%%%%%%%%%%%%%%%%%%%%
\begin{figure*}[t!]
  \centering
  \includegraphics[width=0.9\linewidth]{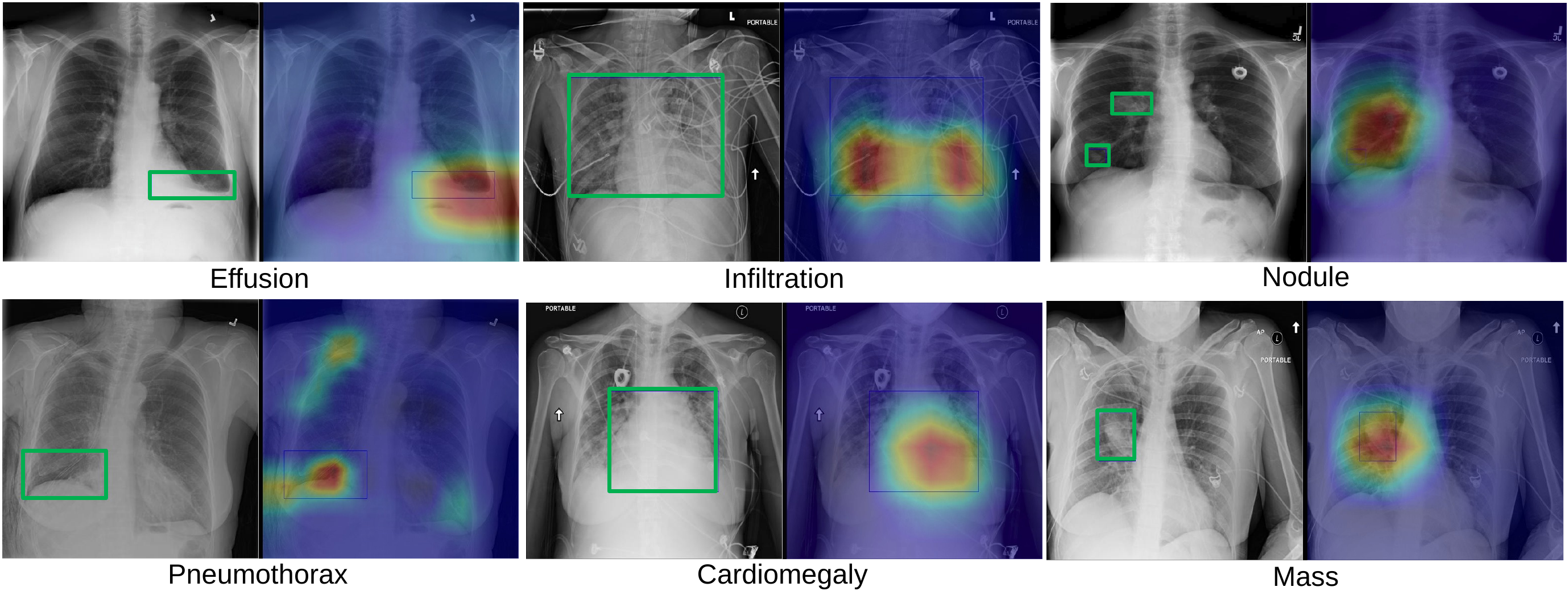}
  \caption{Selected examples of disease localization results (heatmaps overlaid on X-ray images; red color indicates stronger responses) on the ChestXray14 test set using our proposed AGCL framework. The ground-truth bounding boxes are shown in green, and the disease labels are given under each pair of images. }
  \label{fig:results-loc}
\end{figure*}
%%%%%%%%%%%%%%%%%%%%%%%%%%%%%%%%%%%%%%%%%%%%%%%%%%%%%%%%%%%%%%%%%%%%%%%%%%%%%%%%%%%%%%

\textbf{Weakly Supervised Disease Localization:}
We generate bounding boxes from the disease-specific heatmap for each image with corresponding disease, following the benchmark method applied in \cite{Zhou_CVPR2016} and \cite{Wang_CVPR2017}, and evaluate their qualities against ground-truth (GT) bounding boxes annotated by radiologists. A  box is considered as a true positive (TP) if its Intersection of the GT Over the detected Bounding Box area ratio (IoBB~\cite{Wang_CVPR2017}, similar to Area of Precision or Purity) is larger than a threshold $T(IoBB)$. Table. \ref{table-results-loc} shows the comparison of the baseline model, AGL and AGCL-2 (denoted as AGCL in the table.)

Overall, AGCL achieves the best localization results by generating the least number of bounding boxes (3498) but with the most number of true positives (1518), namely the highest precision (0.44). It recalls 73\% of the ground-truth boxes by proposing an average of 3.5 boxes per image. AGCL employs more seed images than AGL by incorporating disease severity level (DSL) based curriculum learning, which improves upon AGL as shown in Table~\ref{table-results-loc}. The relative performance improvements of AGCL are more significant for \textit{Effusion}, \textit{Mass} and \textit{Infiltration}, where more \textit{moderate} and \textit{severe} samples are labeled than other disease types. \textit{Nodule} is often labeled as \textit{small} therefore curriculum learning barely helps. However, visual attention based iterative learning (AGL) outperformed the baseline model even for very difficult diseases such as \textit{Nodule} and \textit{Pneumothorax}. We show some qualitative localization heatmap examples in Fig.~\ref{fig:results-loc}.

\section{Conclusion}

In this paper, we exploit to utilize the NLP-mined disease severity level information from radiology reports to facilitate the curriculum learning for more accurate thoracic disease classification and localization. In addition, an iterative attention-guided refinement framework is developed to further improve the classification and weakly-supervised localization performance. Extensive experimental evaluations on the ChestXray14 database validate the effectiveness on significant performance improvement derived from both the overall framework and each of its components individually. Future work includes formulating structured reports, extracting richer information from the reports such as coarse location of lesions and using follow up studies, and mining common disease patterns, to help develop more precise predictive models. 

\vspace{-1mm}
\subsubsection*{Acknowledgments.} This research was supported by the Intramural Research Program of the National Institutes of Health Clinical Center and by the Ping An Technology Co., Ltd. through a Cooperative Research and Development Agreement. The authors thank NVIDIA for GPU donation.

\bibliographystyle{splncs03}
\bibliography{ref}
% \section*{Appendix: Springer-Author Discount}

\end{document}